%% file: main_paper.tex
\documentclass[sigconf,authorversion]{acmart}

\usepackage{booktabs} 

\makeatletter
\g@addto@macro{\UrlBreaks}{\do\-}
\makeatother

\copyrightyear{2019}
\acmYear{2019}
\setcopyright{acmcopyright}
\acmConference[WSDM '19]{The Twelfth ACM International Conference on Web Search and Data Mining}{February 11--15, 2019}{Melbourne, VIC, Australia}
\acmBooktitle{The Twelfth ACM International Conference on Web Search and Data Mining (WSDM '19), February 11--15, 2019, Melbourne, VIC, Australia}
\acmPrice{15.00}
\acmDOI{10.1145/3289600.3290619}
\acmISBN{978-1-4503-5940-5/19/02}

\begin{document}
\title{TopExNet: Entity-centric~Network~Topic~Exploration~in~News~Streams}
\renewcommand{\shorttitle}{TopExNet: Entity-centric Network Topic Exploration in News Streams}

\author{Andreas Spitz}
\affiliation{%
  \institution{Heidelberg University}
  \streetaddress{Im Neuenheimer Feld 205}
  \postcode{69120}
  \city{Heidelberg} 
  \country{Germany}
}
\email{spitz@informatik.uni-heidelberg.de}

\author{Satya Almasian}
\affiliation{%
  \institution{Heidelberg University}
  \streetaddress{Im Neuenheimer Feld 205}
  \postcode{69120}
  \city{Heidelberg} 
  \country{Germany}
}
\email{almasian@stud.uni-heidelberg.de}

\author{Michael Gertz}
\affiliation{%
  \institution{Heidelberg University}
  \streetaddress{Im Neuenheimer Feld 205}
  \postcode{69120}
  \city{Heidelberg} 
  \country{Germany}
}
\email{gertz@informatik.uni-heidelberg.de}

\input{section_0_abstract}

%
%


\maketitle

\input{section_1_introduction}

\input{section_2_relatedwork}

\input{section_3_theory}

\input{section_4_architecture}

\input{section_5_demonstration}

\input{section_6_summary}

\bibliographystyle{ACM-Reference-Format}
\bibliography{bibliography}

\end{document}

%% file: section_0_abstract.tex
\begin{abstract}
The recent introduction of entity-centric implicit network representations of unstructured text offers novel ways for exploring entity relations in document collections and streams efficiently and interactively. Here, we present TopExNet as a tool for exploring entity-centric network topics in streams of news articles. The application is available as a web service at \url{https://topexnet.ifi.uni-heidelberg.de}.
\end{abstract}

%% file: section_1_introduction.tex
\section{Introduction}

\emph{Keeping up with the news} is a common idiom that is increasingly describing a race that human readers cannot hope to win. Since the publication of news has all but shifted from traditional print media to a rapid stream of online news, we are faced with a constant deluge of news information from the global news cycle. Finding relevant information can be such a daunting task that many users resort to reading nothing but headlines, while news publishers advertise for their articles with prominently displayed reading times of as few minutes as possible. As a result, the larger context is often lost.

An automated aggregation of news is thus clearly beneficial, yet no less daunting from a computational perspective. While much research has been devoted to techniques for finding and linking incidents in news~\cite{DBLP:conf/cikm/FengA07}, such an approach is far from trivial and too restrictive in purely exploratory settings. Intuitively, topic models~\cite{Blei2012} should offer a solid solution to the extraction of relevant topics from collections of documents. However, their performance tends to suffer on large collections of news articles with a multitude of diverse topics, and they are ill-suited for the interactive exploration of documents. Furthermore, topics are usually represented as ranked lists of words, which can be difficult to interpret~\cite{DBLP:conf/nips/ChangBGWB09}.

In this respect, a recent shift in focus towards network-centric representations of documents stands to provide more intuitive and more \emph{visual} access to the complex relations contained in the texts. Examples include the use of concept maps as summaries instead of text snippets~\cite{DBLP:conf/emnlp/FalkeG17}, or the network-centric view on entities as stitching points between interwoven news streams~\cite{DBLP:conf/www/SpitzG18}. Here, we focus on the extraction of topics as network structures of entities and terms~\cite{DBLP:conf/ecir/SpitzG18}, and on how they can be used to explore news.

\noindent
\textbf{Contributions.} 
We present an application that demonstrates how implicit networks 
can be used to discover and expand entity-centric topics in a stream of news articles. By representing entity and term relations as the edges of a network, this approach supports the selection of news outlets and date ranges as additional degrees of freedom, while retaining query speeds that support an interactive use. In contrast to traditional topic models, this results in a more dynamic exploration of topics that can be used in place of aggregation approaches for incidents or articles, such as Google News.

%% file: section_2_relatedwork.tex
\section{Related Work}

Related work covers the areas of topic models and news exploration.

\noindent
\textbf{Topic models.} 
Since the introduction of Latent Dirichlet Allocation~\cite{Blei2003}, numerous variations of topic models have been proposed~\cite{Blei2012}. The majority of these approaches are based on graphical models, which are computationally expensive and ill-suited to interactive use. While efforts have been made to develop more dynamic topic models~\cite{DBLP:conf/icml/BleiL06}, it is not viable to continuously re-compute topics during the interactive exploration of news streams. Furthermore, since traditional topics are fundamentally lists of ranked words that are difficult to visualize, a dynamic exploration of evolving document collections with topic models is problematic. 

Nevertheless, some applications have been presented that support a visual and interactive analysis of topics. One such example are TopicNets~\cite{DBLP:journals/tist/GretarssonOBHANS12}, which allow the user to view document contents within the larger scope of overarching topics. Similarly, word network topics are designed for the discovery of topic relations in short texts~\cite{DBLP:journals/kais/ZuoZX16}. Unlike our approach, these applications lack a focus on entities as anchors of event descriptions in news texts.

\begin{figure*}[t]
  \centering
  \includegraphics[width=0.98\textwidth]{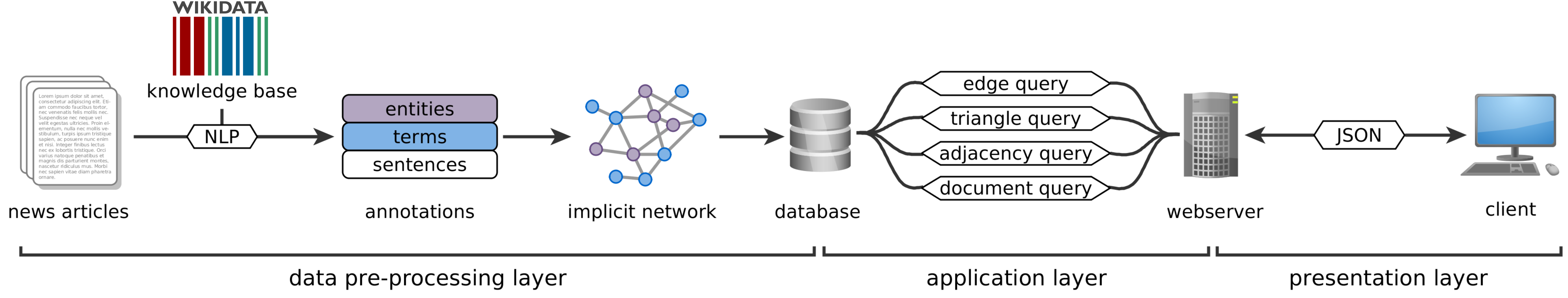}%
  \Description%
  
  \vspace*{-0.3cm}
  
  \caption{Schematic view of the architecture for extracting and querying implicit network topics from streams of news articles.}
  \label{fig:architecture}
  
 \vspace*{-0.2cm}  
  
\end{figure*}

\noindent
\textbf{News exploration.} 
In contrast to topic models, some tools specifically support the exploration of news streams with a focus on entities, such as STICS~\cite{DBLP:conf/sigir/HoffartMW14} or EventRegistry~\cite{DBLP:conf/www/LebanFBG14}. NewsStand takes a different approach by clustering news articles geographically~\cite{DBLP:conf/gis/TeitlerLPSSS08}. Further approaches include the monitoring of multilingual European news~\cite{DBLP:conf/www/AtkinsonG09} and the extraction of semantic word clouds with significance analysis to obtain a quick overview over current news~\cite{DBLP:conf/ecir/SchubertSZG18}. 

However, none of these tools include an exploration of topics. To fill this gap, we rely on an implicit network representation of text as used in EVELIN~\cite{DBLP:conf/www/SpitzAG17}, which was designed for static document collections and with single relations between entities in mind. We improve upon this concept by adding an exploration of the more complex graph structures that are inherent to network topics.

%% file: section_3_theory.tex
\section{Theoretical Background}
\label{sec:theory}

We give a brief overview of the construction of network topics. For a more detailed description, we refer to our previous work~\citep{DBLP:conf/ecir/SpitzG18}.

\subsection{Implicit Networks}

Conceptually, an implicit network can be viewed as a word cooccurrence graph in which (1) entities are linked to a knowledge base, (2) long-range cooccurrences beyond sentence boundaries are considered, and (3) edge weights are derived from cooccurrence distances instead of counts. Individual edges are then aggregated over all documents to create a network representation, which may include sentences and documents as nodes. In a streaming setting, it can be sensible to partially aggregate edges in preprocessing~\cite{DBLP:conf/www/SpitzG18}.

In the following, we consider the network to be a list of edge tuples $e=\langle v, w, t, out, d, \delta\rangle$, where $v$ and $w$ are two nodes (entities or terms), $t$ is the publication date, $out$ is the news outlet, $d$ is the document, and $\delta$ is the minimum textual distance between the two nodes in the document (measured in sentences). To support efficient queries over varying date ranges and selections of news outlets, we partially aggregate edges to at most one edge per entity pair, publication day, and outlet. Node statistics such as the occurrences in documents are partially aggregated in a similar fashion.

\subsection{Network Topic Extraction}

Motivated by the important role that entities play in news events, important edges between entities are considered as \emph{topic seeds}, around which a shell of descriptive terms is constructed. If terms are ranked according to their importance for an edge, each such subgraph can be regarded as a ranked list of terms, similar to traditional topics, yet more visual. 
To discover the most important seed edges and select relevant terms, we thus require edge weights to rank the relations. By including the date range, we use a weighting scheme with three components that are combined as the harmonic mean. To derive the weight of an edge $e=(v,w)$ between nodes $v$ and $w$ in a date range $t=(t_1,t_2)$, let the score $\omega$ be
$$
\omega(e, t)=3\left[ \frac{|D_v \cup D_w|}{|D_{e}|} 
+ \frac{t_1 - t_2}{|T_e|} 
+ \frac{D_{max}}{\Delta_e}  
\right]^{-1}
$$
where $D_v$, $D_w$, and $D_{e}$ denote the sets of documents in which $v$, $w$, and $e$ occur, $T_e$ is the set of days on which $e$ is mentioned, $D_{max}$ is the maximum number of documents any edge is mentioned in, and 
$\Delta_e = \sum_{e}\exp(-\delta_e)$ is the sum of decaying reciprocal distances. 

Based on this scoring function, it is then a simple matter to select edges that correspond to important cooccurrences. In the application, entity edges can be instantiated either by selecting the globally highest ranked edges for a time interval and set of outlets, or by directly specifying pairs of entities that are of interest to the user. In either case, descriptive terms are added by ranking them according to their importance for the entities of each edge, thereby inducing triangular subgraphs. If edges of distinct subgraphs overlap in an entity, they can be merged into a larger topic subgraph.

%% file: section_4_architecture.tex
\section{System Architecture}
\label{sec:architecture}

In the following, we describe the system architecture for data preprocessing and network topic extraction as shown in Figure~\ref{fig:architecture}.

\begin{figure*}[t]
  \centering
  \includegraphics[width=1.0\textwidth]{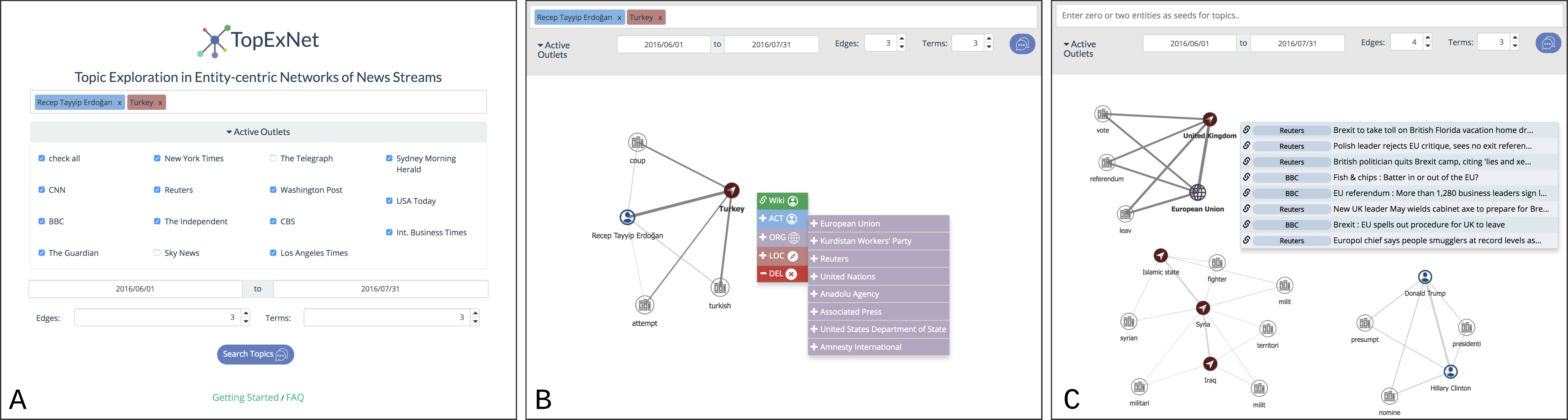}%
  \Description%
  
  \vspace*{-0.3cm}
  
  \caption{Overview of TopExNet's user interface. A: initial search page with selectable parameters. B: result of a targeted entity edge query with an entity exploration menu. C: result of a global edge ranking query with an article recommendation menu.}
  \label{fig:screenshot}
  
 \vspace*{-0.3cm}  
  
\end{figure*}

\subsection{Data Preprocessing}

Implicit networks can be extracted from any document that is annotated for entities. Since the cooccurrences in each document correspond to a small network and networks are additive, document streams can be iteratively composed into a larger network that represents the entire stream history as the documents arrive. 

Document preprocessing includes part-of-speech and sentence tagging, named entity recognition and linking, and entity classification. Stanford CoreNLP~\cite{DBLP:conf/acl/ManningSBFBM14} is used for sentence splitting and part-of-speech tagging. For the recognition and disambiguation of named entities to Wikidata IDs, we use Ambiverse\footnote{\url{https://www.ambiverse.com/}}. To identify named entities of type actor, location, and organization, we map Wikidata IDs to YAGO3 entities~\cite{DBLP:conf/semweb/RebeleSHBKW16} and classify them according to the YAGO hierarchy by using class \texttt{wordnet\_person\_100007846} for actors, class \texttt{wordnet\_social\_group\_107950920} for organizations, and \texttt{yagoGeoEntity} for locations. Finally, all entities are augmented with Wikidata descriptions. Remaining tokens that are at least four characters long constitute the set of terms and are stemmed with a Porter stemmer~\cite{porter1980}.
The implicit network is constructed from the annotated data as outlined in Section~\ref{sec:theory}. For entity cooccurrences, we set the window size to 5 sentences, and use intra-sentence occurrences for terms and entities.

\subsection{Application Layer} 
While an in-memory representation of the data is possible and supports fast query processing, it does not scale arbitrarily and is not feasible for a long-running non-commercial demonstration. Therefore, we use a Core i7 with 32GB main memory and an SSD drive as demonstration server. The network is stored in a MongoDB, with separate collections for entities, terms, edges between entities, and edges between entities and terms. Entities are enriched with Wikidata information to provide entity descriptions at query time. Based on input strings, a text index on the English canonical label is used to compile a list of entity suggestions for the user. We rank entity suggestions by the text match score and break ties by the number of occurrences. All edges are partially aggregated at a granularity level of days to speed up subsequent aggregations at query time. Node occurrence information is stored in a similar aggregation to retrieve individual occurrence counts in documents.

The interactive topic extraction methods described in Section~\ref{sec:theory} are implemented in Java and enable query processing speeds in the order of a few seconds for all but the most highly connected entities. Like most other complex networks, implicit networks have a long-tailed degree distribution, which translates to the presence of few highly connected hubs in practice. While queries on the network generally benefit from the overall sparseness, hubs may cause longer query response times for incident edges, especially for large date ranges. However, due to their small number, this problem can be addressed by caching results in the database, which ameliorates the effect over time. Specifically, we use a separate collection for caching the results of individual triangular term expansion queries that then serve as building blocks of later queries.

While topic extraction queries can be parallelized by edge, we only allocate one thread per query to serve queries from multiple users simultaneously. To avoid system overload in the case of multiple users, we use an anonymized mapping of queries to browser fingerprints and limit the number of active queries per user.

\subsection{Presentation Layer}
The web interface is implemented in HTML and JavaScript, and accepts user input to extract suitable topics and visualize the output as graphs. For entity input and for sending queries to the application layer, we use jQuery. The Bootstrap libraries\footnote{\url{http://getbootstrap.com}} and Mustache web templates enable the interactive layout. To recognize and classify input entities, we use the tags-input and typeahead libraries of Bootstrap, which we extend by adding the required functionality for the color coding of entities. The interactive visualization of the topic networks and the menus is handled by the vis.js JavaScript library\footnote{\url{http://visjs.org/}}. Graphs are visualized with a force-directed layout.

The web server itself uses the Java Spark micro framework\footnote{\url{http://sparkjava.com}} and is directly integrated with the application layer. Communication between user interface and server is built on AJAX and uses JSON for transmitting data in both directions (i.e, input query entities, output graph data). Examples of the interface are shown in Figure~\ref{fig:screenshot}, based on which we discuss the functionality in the following.

%% file: section_5_demonstration.tex
\section{Functionality and Demonstration}

We briefly describe the data used in the demonstration, before discussing TopExNet's functionality and usage scenarios.

\subsection{News Network Data}

As data for the presentation, we collect articles from the RSS feeds of international news outlets with a focus on politics. The content is extracted with manually created rules to include multi-page articles and avoid the drawbacks of boilerplate removal. 
Specifically, we use articles from 14 English speaking news outlets located in the U.S. (CNN, LA Times, NY Times, USA Today, CBS News, The Washington Post, IBTimes), Great Britain (BBC, The Independent, Reuters, SkyNews, The Telegraph, The Guardian), and Australia (Sydney Morning Herald) during the period from June 1 to November 30, 2016. We remove articles that have less than 200 or over $20,000$ characters (due to limitations in the NER step) or more than 100 disambiguated entities per article (i.e., lists). The final collection contains $127,485$ articles with a total of 5.4M sentences.
After preprocessing as described in Section~\ref{sec:architecture}, the resulting network has 27.7K locations, 72.0K actors, 19.6K organizations, and 322K terms, which are connected by 26.8M partially aggregated edges.

\subsection{Input Parameters}

User input for queries to the data is based on the four parameters \emph{entities}, \emph{date range}, the \emph{number of edges} and the \emph{number of terms}. Additionally, a subset of news outlets can be selected.

\emph{Entities} are entered as input by selecting them from a list of suggestions that is generated from one or several strings provided by the user. Entity suggestions are automatically linked to network nodes upon selection. A \emph{date range} is chosen by using a date range picker with a granularity of days, and is limited to the publication time frame of articles in the stored stream. The \emph{number of edges} can be used to set the number of seed edges when global edge ranking serves as a starting point. Similarly, the \emph{number of terms} that are extracted for each seed edge can be adjusted. An overview over the initial input screen is shown in Figure~\ref{fig:screenshot}A.

\subsection{Exploration Approaches}

TopExNet offers four primary modes of exploring network topics and the underlying news stream as we describe in the following.

\noindent
\textbf{Global edge ranking.} 
The first primary use-case is the automatic extraction of seed edges from the network. If the user specifies a date range but no input entities, then the output is a global ranking of all entity edges for the specified date interval and news outlets. The highest ranked edges are selected as topic seeds, merged if they share some of their nodes, and expanded by adding descriptive terms. An example of the output for three edges and three terms per edge is shown in Figure~\ref{fig:screenshot}C. For a larger number of edges and a limited time frame, this serves as a birds-eye view on current news.

\noindent
\textbf{Targeted entity exploration.} 
In contrast to the global ranking of edges, which focusses on the topics surrounding the entities that are the overall most central during the selected time frame, the user may also focus on specific entities. When supplied with two entities as query input, TopExNet adds descriptive terms only to the edge between the two provided entities. An example is shown in Figure~\ref{fig:screenshot}B. While such individual seed edges naturally generate smaller topic networks, they serve as selectable starting points and can be further expanded by adjacency exploration.

\noindent
\textbf{Topic network exploration.} 
The network topics from the above two cases support further exploration. In addition to the obvious tuning by increasing or decreasing the number of seed edges or descriptive terms, the user may also expand the displayed network. When selecting any entity in the network, the user is given the choice to include highest ranked adjacent nodes. Here, we rely on the entity ranking method that was introduced for EVELIN~\citep{DBLP:conf/www/SpitzAG17}, and adapt it to semi-aggregated edges. A visualization of the process is shown in Figure~\ref{fig:screenshot}B. Once additional entity edges are added, descriptive terms may be included by clicking on edges or the canvas and selecting the option to add terms. Likewise, nodes that are not of interest can be deleted, or information about entities can be obtained by opening the corresponding Wikidata pages.

All three of the above approaches support a facetted and contrastive analysis of network topics. By viewing time slices or topics for subsets of outlets in parallel windows, the user can compare network topics and their evolution between different sources.

\noindent
\textbf{Article recommendation.} 
Finally, once the user has identified topics of interest, TopExNet can recommend suitable news articles that describe the selection in-depth. When \emph{multiple} entities and/or terms are selected, a right-click opens a menu of article recommendations. Each recommendation links to an original news article that is relevant to the selected nodes. Thus, the exploration and identification of network topics can serve as an entry point for the focused reading or analysis of related news articles.

%

%% file: section_6_summary.tex
\section{Summary and Outlook}

In this demonstration, we presented TopExNet as a web-based application for the exploration of network topics in news streams. By leveraging implicit entity network representations of the underlying document stream, we demonstrated the feasibility of exploring entity-centric topics interactively, even for large document collections and in a streaming setting. 
On the technical side, the partial aggregation of entity and term cooccurrence edges allows the efficient retrieval of both seed edges and descriptive terms from the data, without the costly requirement of re-computing topics for the entire corpus due to changing parameters, thus making the extraction of topics truly dynamic. 
On the application side, the network representation of topics aggregates content without the need to display the content of potentially proprietary news articles to the user, and thus serves as a valuable alternative to existing news aggregation and summarization approaches in industrial settings.

\noindent
\textbf{Future work.} 
For this demonstration, we presented TopExNet as a standalone application. However, the underlying network structure is similar to the data representation used for EVELIN~\citep{DBLP:conf/www/SpitzAG17}, meaning that topic exploration can be integrated seamlessly. To ensure scalability with an increasing number of news outlets, we are considering a replication of the data in clustered database servers to benefit from the parallel nature of individual edge queries.